\documentclass[a4paper]{article}

\usepackage{multirow}
\usepackage{url}
\usepackage{hyperref}
\usepackage{diagbox}
\usepackage{color}

\usepackage{float}
\newcommand\figcaption{\def\@captype{figure}\caption} 
\newcommand\tabcaption{\def\@captype{table}\caption} 

\usepackage{INTERSPEECH2020}

\title{TMT: A Transformer-based Modal Translator for Improving Multimodal Sequence Representations in Audio Visual Scene-aware Dialog}
\name{Wubo Li, Dongwei Jiang, Wei Zou, Xiangang Li}
\address{
  Didi Chuxing, Beijing, China}
\email{\{liwubo, jiangdongwei, zouwei, lixiangang\}@didiglobal.com}

\begin{document}

\maketitle
\begin{abstract}
Audio Visual Scene-aware Dialog (AVSD) is a task to generate responses when discussing about a given video. The previous state-of-the-art model shows superior performance for this task using Transformer-based architecture. However, there remain some limitations in learning better representation of modalities. Inspired by Neural Machine Translation (NMT), we propose the Transformer-based Modal Translator (TMT) to learn the representations of the source modal sequence by translating the source modal sequence to the related target modal sequence in a supervised manner. Based on Multimodal Transformer Networks (MTN), we apply TMT to video and dialog, proposing MTN-TMT for the video-grounded dialog system. On the AVSD track of the Dialog System Technology Challenge 7, MTN-TMT outperforms the MTN and other submission models in both Video and Text task and Text Only task. Compared with MTN, MTN-TMT improves all metrics, especially, achieving relative improvement up to $14.1\%$ on CIDEr.
\end{abstract}

\section{Introduction}
\label{sec:intro}
Recently, a considerable amount of literature has been published on multimodal learning of visual and language. These studies concern the tasks of media description and question answering, such as video caption \cite{krishna2017dense, xu2016msr-vtt, chen2011collecting}, video question answering \cite{xu2017video, tapaswi2016movieqa, lei2019tvqa} and video dialog \cite{alamri2019audio}. Audio Visual Scene-aware Dialog (AVSD) is the task of generating a response for a question with a given scene, video, audio, and the history of previous turns in the dialog.  The AVSD is considered one of the most challenging tasks because the system needs to recognize the history of dialog along with the visual and acoustic data for accurately answering the question \cite{Lee2020DSTC8}. 

One of the challenges in the AVSD task: acquiring accurate representations of the multiple modalities. To tackle this, some previous studies focus on pre-training a modal extractor. In \cite{DBLP:conf/iccv/HoriHLZHHMS17,sanabria2019cmu, kumar2019context}, features are extracted from the fix-frame video and audio by pre-trained I3D \cite{carreira2017quo}, VGG \cite{hershey2017cnn} and 3D ResNeXt \cite{hara2018can}. Sanabria et al. \cite{sanabria2019cmu} use 2000 hours of how-to videos \cite{sanabriahow2} to pre-train a video modal extractor and treat the video-grounded dialog task as a video caption task to generate answers. 
In the textual question answering task, lots of studies use questions to figure out useful information from documents \cite{kumar2016ask, seo2016bidirectional}. This method can be transferred to the multimodal dialog system as well. The work in \cite{yeh2019reactive} uses a simple but effective $1\times1$ convolution to fuse multimodal features and proposes a multi-stage fusion mechanism to thoroughly understand the question. Inspired by FiLM, FiLM Attention Hierarchical Recurrent Encoder-Decoder (FA-HERD) \cite{nguyen2019film} learns the representation of multiple modalities by using FiLM blocks to condition video and audio on the embeddings of the current question. \cite{le2019multimodal} proposed query-aware auto-encoder to obtain accurate representations from visual and audio modalities. 

Recently, the evidence \cite{tsai2019multimodal} reports that translating cross-modalities is beneficial to capture correlated signals across asynchronous modalities. Along with this motivation, we try to learn the representation of modality by the consistency of related modalities.

In this paper, we propose the Transformer-based Modal Translator (TMT) for learning representation of modalities. In the TMT, the source modal sequence is translated to other related modal sequences in a supervised learning manner. The key to our method is that we effectively make use of correlative multiple modalities to express source modality. Based on MTN, we apply TMT to video and dialog, proposing MTN-TMT for the video-grounded dialog system. We evaluated MTN-TMT on the AVSD track of the $7^{th}$ Dialog System Technology Challenge, which generates dialog responses considering multiple modalities. MTN-TMT outperforms the MTN and other submission models in both Video and Text task and Text Only task. Compared with MTN, MTN-TMT improves all metrics, especially, achieving relative improvement up to $14.1\%$ on CIDEr.

\section{Related Work}
\label{sec:related-work}
The work relevant to ours is MTN proposed by Le et al. \cite{le2019multimodal}, which is a Transformer-based encoder-decoder framework that has several attention blocks to incorporate multiple modalities such as video, audio, and text. The MTN is composed of three major components: encoder, decoder, and query-aware auto-encoder. In the encoder, the text sequence and video features are mapped to a sequence of continuous representations $ z=(z_1,...,z_n) \in \mathbb{R}^{d} $, $ f_a $ and $ f_v $, where $d$ is the dimension of word embeddings. The aim of query-aware auto-encoder is to learn the representations of non-textual modalities that are relevant to the query, such as audio and visual. Decoder consists of a stack of Transformer layers to fuse all modalities and generate an answer $y=(y_1,...,y_m)$ including an end-of-sentence token (eos). We refer the reader to \cite{le2019multimodal} for a more detailed explanation of the MTN. One of the contributions of MTN is that replacing RNN with Transformer enhances the context-dependence of modality. Another contribution is that they propose a query-aware attention encoder to learn the representations of non-text modalities. Compared with query-aware auto-encoder, TMT learns representations of modalities by other related modalities instead of query.

Multimodal Transformer (MulT) \cite{tsai2019multimodal} is another work related to ours which uses decoder Transformer in NMT to align cross-modalities. They hypothesize a good way to fuse cross-modal information is providing a latent adaptation across modalities. For human multimodal sentiment analysis and human emotion analysis tasks, they use six Transformer-based translators to align each pair of modalities among visual, audio and language. Similar to MulT, TMT uses a machine-translation-like method to translate the source modality to the related target modality. In addition, we treat target sequence modality as a label for supervised learning.

\section{Our Approach}
\label{sec:approach}

\subsection{Transformer-based modal translator}
\label{ssec:TMT}
\begin{figure}[ht]
  \centering
  \centerline{\includegraphics[scale=0.35]{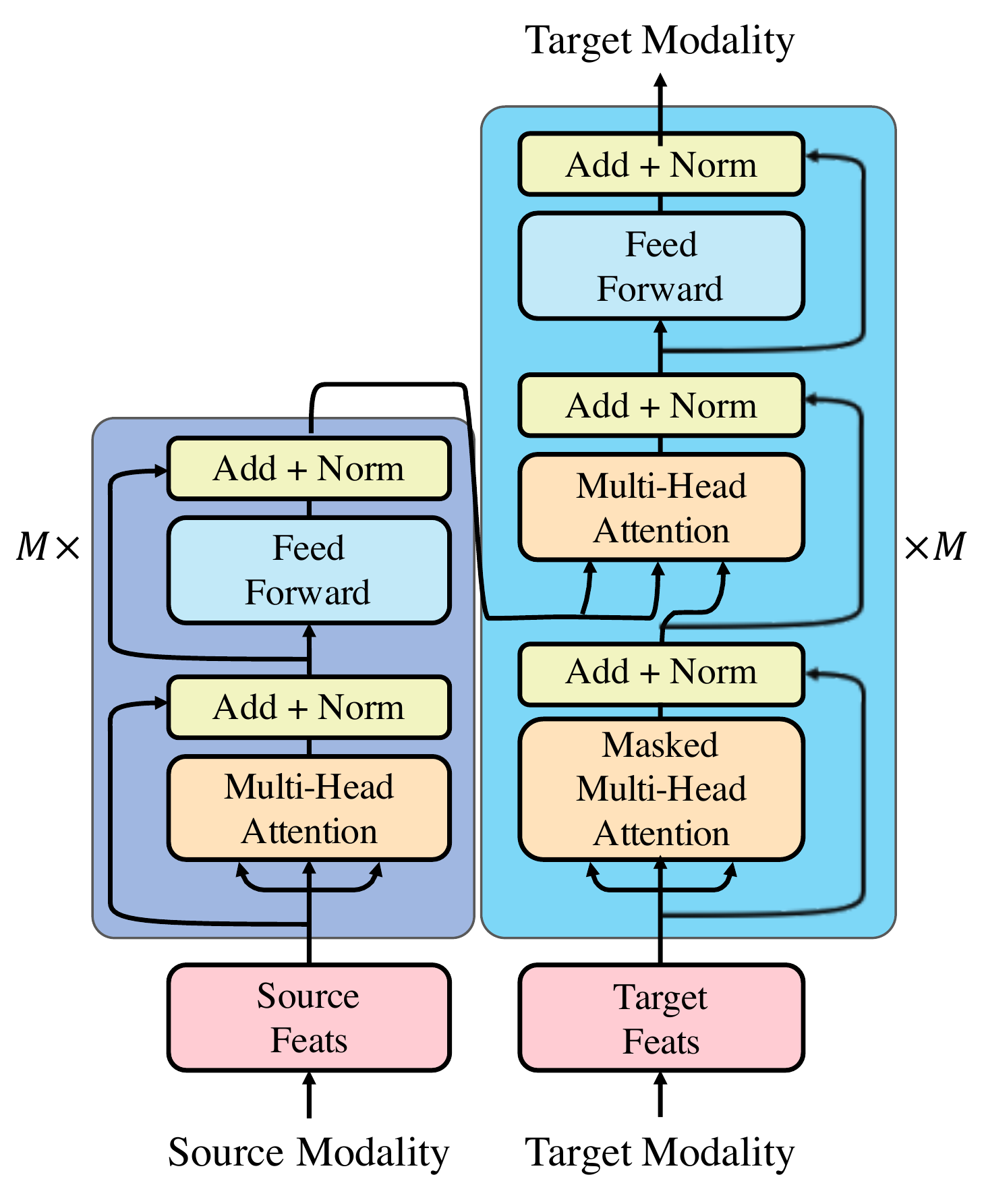}}
  \vspace{-0.90cm}
  \centerline{}\medskip
\caption{Overall Architecture of TMT.}
\label{TMT}
\end{figure}

Motivated by that translating related modalities could deeply capture the relevance between modalities, we introduce Transformer-based Modal Translator (TMT) to translate the source modal sequence to the related target modal sequence to learn better source modal sequence. As shown in Figure \ref{TMT}, TMT is made up of multi-layer Transformer encoder (left) and Transformer decoder (right). In this work, TMT with $M$ layers indicates that the depth of Transformer encoder and Transformer decoder are both $M$. Given the source modal sequence $X = {\{x_1, x_2,..., x_T\}}$ and target modal sequence $Y={\{y_1, y_2,..., y_N\}}$, where $T$ and $N$ is the length of the source modality and target modality respectively. TMT mainly consists of two steps. At the first step, a Transformer encoder is used to encode the source modal sequence. Then, a Transformer decoder is used to translate the source modal sequence to the target modal sequence. In order to learn the mapping between the source modal sequence and the target modal sequence, we use supervised learning to fit the output sequence $\widehat{Y}=\{\widehat{y_1}, \widehat{y_2},..., \widehat{y_N}\}$ of the TMT and the target modal sequence $Y$. When the target modality is text, we employ cross-entropy as the loss function. As for dense modalities, such as speech and images, we use the $L1$ loss or similarity loss \cite{kovaleva-etal-2018-similarity}.

\subsection{MTN-TMT}
\label{ssec:mtn-TMT}

\begin{figure*}[th]
  \centering
  \centerline{\includegraphics[scale=0.56]{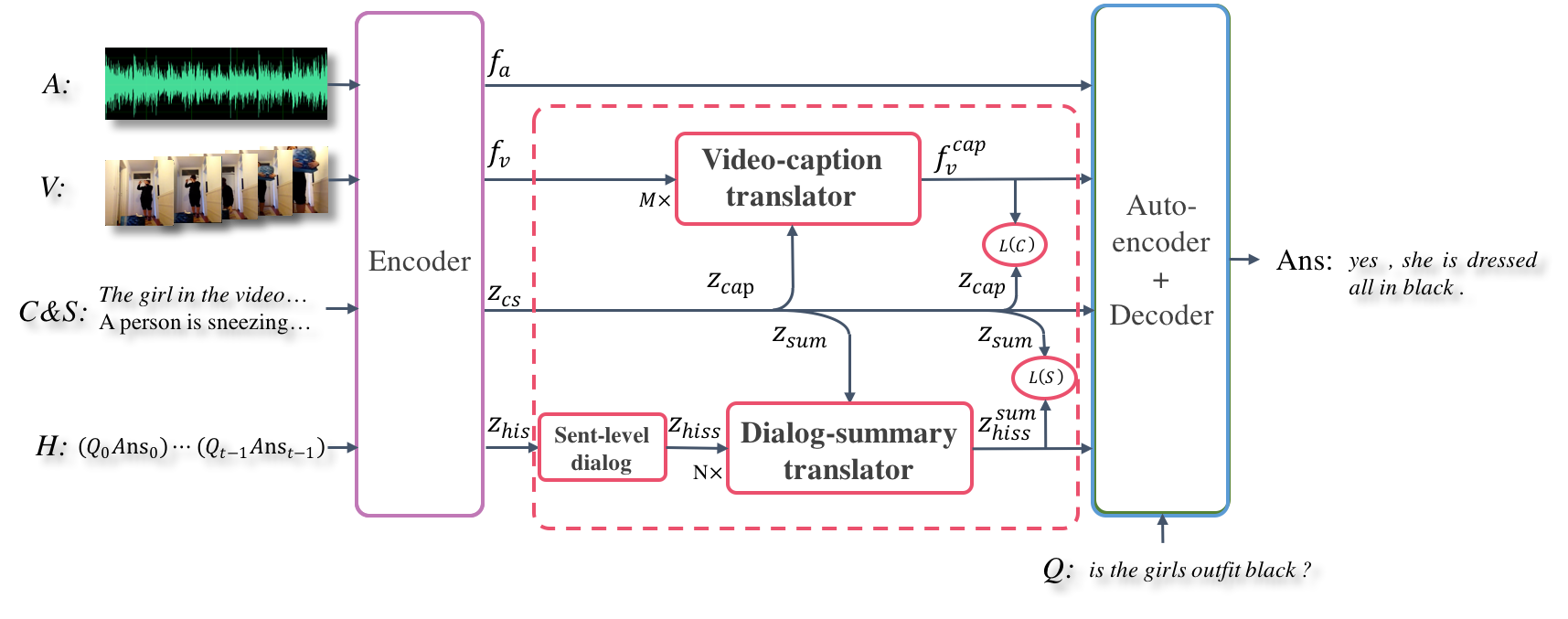}}
  \vspace{-1.1cm}
  \hspace{-10.0cm}
  \centerline{}\medskip
\caption{Overall Architecture of MTN-TMT. Video-caption translator and dialog-summary translator are highlighted by red dotted rectangle. The $A$, $V$,  $C$, $S$, $H$, $Q$ and $Ans$ stand for audio, video, caption, summary, dialog history, question, and answer, respectively.}
\label{MTN-TMT-new}
\end{figure*}

Recently, \cite{le2019multimodal} proposed Multimodal Transformer Networks (MTN), which is the state-of-the-art system for DSTC7-AVSD. Based on MTN, we propose a novel MTN-TMT applying TMT to video and dialog. Figure \ref{MTN-TMT-new} shows the overall architecture of MTN-TMT, which includes encoder, auto-encoder, decoder, video-caption translator and dialog-summary translator. We follow the pre-processing and structure of MTN \cite{le2019multimodal}, except adding TMT-based video-caption translator and dialog-summary translator after encoder for learning representations of video and dialog. Next, we will briefly introduce the video-caption translator, dialog-summary translator and loss function.


\subsubsection{Video-caption translator } \label{subsec:vc-translator}
We consider caption as a relevant modality to video which summarizes the video's content. To enhance the representations of video, we propose a TMT-based video-caption translator to translate the video to caption. Figure \ref{MTN-TMT-new} shows the data flow from encoder to the video-caption translator. We feed visual representations $ f_v $ as source modality and tokenized caption $z_{cap}$ as target modality into video-caption translator. Video-caption translator consists of $M$ layers of TMT. The output of the video-caption translator are caption-related visual representations $f^{cap}_v$.

\subsubsection{Dialog-summary translator} \label{subsec:ds-translator}
Although dialog history usually contains a wealth of details, it remains a challenge to capture useful information from a long dialog history. Given that dialog history and summary are describing the same content in different ways, we introduce a dialog-summary translator to translate the dialog history into a summary to enhance representations of dialog history.
Dialog-summary translator consists of $N$ layers of TMT. As shown in Figure \ref{MTN-TMT-new}, the input modalities of the dialog-summary translator are sentence-level dialog history $z_{hiss}$ and summary representations $z_{sum}$. Inspired by work in \cite{Hori_2019}, before translating, we use hierarchical Transformer layer to preprocess tokenized dialog history sequences $z_{his}$ to get the sentence-level dialog history $z_{hiss}$. The hierarchical Transformer consists of  $N$ Transformer encoder layer \cite{vaswani2017attention}. The sentence-level dialog history $z_{hiss}$ are outputs of the eos tokens in dialog history. Then summary representations $z_{sum}$ and the sentence-level dialog history $z_{hiss}$ are fed into dialog-summary translator to generate summary-related dialog representations $z^{sum}_{hiss}$.

\subsubsection{Loss} \label{subsec:loss}
Given the dialog history ($H$), question ($Q$), video features (video $V$ and audio $A$), video caption ($C$) and dialog summary ($S$), we use the log-likelihood as the objective function for target sequences answer ($Ans$), video caption ($C$) and dialog summary ($S$).  The log-likelihood function consists of three parts respectively:
\begin{normalsize}
\begin{equation} \label{eqn1}
  \begin{split}
  L=&L(Ans) +\alpha L(C) + \beta L(S)    \\
  =&logP(Ans| H, Q, V, A, C, S) + \alpha logP(C| V)  + \\
  & \beta logP(S| H).
  \end{split}
\end{equation}
\end{normalsize}
where $\alpha$ and $\beta$ are learning weights for the video-caption translator and dialog-summary translator in loss function.

\section{Experiments}
\label{sec:experiment}

\subsection{Datasets}
\label{ssec:data}

\begin{table}[b]
\setcounter{table}{0}
\centering	
        \caption{DSTC7 Audio Visual Scene-aware Dialog Dataset}
        \setlength{\baselineskip}{1.5em}
	\setlength{\tabcolsep}{1.2mm}
	\begin{tabular}{lccc}
    \hline
	& Training & Dev & Test  \\ \hline
	num of Dialogs     & 7,659          & 1,787               & 1,710          \\ 
	num of Turns       & 153,180        & 35,740              & 13,490         \\ 
	num of Words       & 1,450,754      & 339,006             & 110,252        \\ \hline
	\end{tabular}
	\label{AVSD_data}
	\end{table}
	
We evaluated MTN-TMT on the Audio Visual Scene-aware Dialog track of the $7^{th}$ Dialog System Technology Challenge (DSTC7-AVSD) proposed on \cite{Hori_2019}. This dataset consists of Q\&A conversations regarding short videos obtained from two Amazon Mechanical Turk (AMT) workers, who discuss events in a video. In each dialog, one of the workers takes the role of an answerer who had already watched the video. The answerer replies to questions asked by another AMT worker, the questioner. The dataset contains 9,848 videos taken from CHARADES, which is a multi-action dataset including 157 action categories \cite{Sigurdsson_2016}.
\begin{table}[t]
	\setcounter{table}{1}
	\setlength{\belowcaptionskip}{-0.5cm}
	\centering
	\caption{Automatic evaluation metrics on DSTC7-AVSD. The VCT and DST stand for  video-caption translator and dialog-summary translator, respectively.}
	\setlength{\baselineskip}{1.5em}
	\setlength{\tabcolsep}{1mm}
	\normalsize
	\begin{tabular}{lcccc}
   \toprule
          Description&	 BLEU-4&	 	METEOR& 		ROUGE-L& 	CIDEr\\  \hline       
	\multicolumn{5}{c}{Text Only} \\\hline
	 MTN \cite{le2019multimodal}& 0.307&	0.283&	0.553&	0.973\\ 
         MTN-TMT& 	\textbf{0.346}&	\textbf{0.289}&	\textbf{0.571}&	\textbf{1.107}\\ \hline
         \multicolumn{5}{c}{Video and Text} \\   \hline
	MTN\cite{le2019multimodal}&	0.392& 0.278& 0.571& 1.128\\
	MTN+VCT&	0.379&	0.286&	0.578&	1.146\\
	MTN+DST&	0.398&	0.291&	0.583&	1.176\\
	MTN-TMT&\textbf{0.402}&	\textbf{0.293}&	\textbf{0.587}&	\textbf{1.190}\\   \toprule
	\end{tabular}
        \label{avsd_result}
	\end{table}

We used 7,659 dialogs for the training set and 1787 dialogs for dev set, and evaluated our model on the official test set which contains 1710 dialogs \cite{Hori_2019}. Table \ref{AVSD_data} summarizes the dataset.

\subsection{Metrics}
\label{ssec:metrics}	
In the DSTC7-AVSD, the metrics are commonly used in natural language process tasks, such as BLEU \cite{papineni2002bleu}, METEOR \cite{denkowski2014meteor}, ROUGE-L \cite{lin2004automatic} and CIDEr \cite{vedantam2015cider}. Note that higher scores are better. The scores of the metrics are obtained by the toolkit provided by the organizer.

\subsection{Experiment Settings}
\label{ssec:experiment_settings}

Except for the hyper-parameters mentioned below, we follow the training settings of MTN \cite {le2019multimodal} in DSTC7-AVSD. We use dropout \cite{JMLR:v15:srivastava14a} with keeping probability 0.5, and warm up the schedule with ${warmup\_step}$ 13000 for modifying the learning rate  \cite{vaswani2017attention}. We adopt the Adam optimizer \cite{kingma2014adam} with ${\beta_1 = 0.9}$, ${\beta_2 = 0.98}$, and ${\epsilon=10^{-9}}$. In all experiments, we select a batch size of 32 and train models up to 35 epochs. While training MTN-TMT, we use the grid search method among hyperparameters $\alpha$ and $\beta$ on dev set. For all models, we evaluated on the test set of the DSTC7-AVSD by loading models with the lowest perplexity on the dev set. We have carried out multiple experiments, and the reported metrics of each model are the average of three experiments. 

\begin{table}[t]
	\setcounter{table}{2}
	\setlength{\belowcaptionskip}{-0.5cm}
	\centering
	\caption{An ablation study on the influence of supervised learning on TMT. WD indicates learning weights decay mentioned in section \ref{ssec:results}}
	\setlength{\baselineskip}{1.5em}
	\setlength{\tabcolsep}{1mm}
	\normalsize
	\begin{tabular}{lcccc}
   \toprule
	Description & BLEU-4 & METEOR & ROUGE-L & CIDEr \\  \hline
	 MTN \cite{Hori_2019} &	0.392& 0.278& 0.571& 1.128\\
	 MTN-TMT (WD)&  0.375&	0.286&	0.574&	1.135\\ 
         MTN-TMT&	\textbf{0.393}& 	\textbf{0.296}&	\textbf{0.587}&	\textbf{1.169} \\ 
   \toprule
	\end{tabular}
        \label{decay_weight}
	\end{table}

We evaluate MTN-TMT model in DSTC7-AVSD 's Video and Text and Text Only tasks \cite{Hori_2019}. In Video and Text task, we use multiple modal input, including dialog history, audio, visual, caption and summary. Text Only task allows only dialog history and caption as input. Compared with the caption, we consider that summary is more relevant to dialog history. As a result, in the Text Only task, we add the summary as an extra input. 

\subsection{Results}
\label{ssec:results}
\textbf{Text Only task.} We first evaluate MTN-TMT in Text Only task. The upper part of the Table \ref{avsd_result} shows the result of MTN and MTN-TMT. For Text Only task, we only add dialog-summary translator to MTN and note it as MTN-TMT. We observe MTN-TMT outperforms MTN, where BLEU-4, METEOR, ROUGE-L, and CIDEr are improved by 0.039, 0.006, 0.018 and 0.134.  \\

\begin{table*}[th]
	\setcounter{table}{4}
	\setlength{\belowcaptionskip}{-1cm}
	\centering
	\caption{Automatic evaluation metrics of the submission models to the DSTC7-AVSD. Note that the evaluation of HAN on one reference is not available.}
	\setlength{\baselineskip}{1.5em}
	\setlength{\tabcolsep}{3mm}
	\normalsize
	\begin{tabular}{lcccccccc}
   \toprule
         \multicolumn{1}{l}{Description}& \multicolumn{4}{c}{Official 1 reference} & \multicolumn{4}{c}{Official 6 references} \\  \cline{2-9}
          &	 BLEU-4&	 	METEOR& 		ROUGE-L& 	CIDEr& 	 BLEU-4& 	METEOR& 	ROUGE-L& 	CIDEr\\  \hline       
	\multicolumn{9}{c}{Text Only} \\\hline
	 HAN \cite{sanabria2019cmu}&	-&	-&	-&	-&	\textbf{0.376}&	0.264&	0.554&	1.076 \\
	 MTN \cite{le2019multimodal}&	0.115&	0.173&	0.358&	1.141&	0.307&	0.283&	0.553&	0.973 \\ 
         MTN-TMT (Ours)& 	\textbf{0.131}&	\textbf{0.174}&	\textbf{0.368}&	\textbf{1.302}&	0.346&	\textbf{0.289}&	\textbf{0.571}&	\textbf{1.107}\\ \hline
         \multicolumn{9}{c}{Video and Text} \\   \hline
	HAN \cite{sanabria2019cmu}&		-&	-&	-&	-&	0.394&	0.267&	0.563&	1.094\\
	MTN \cite{le2019multimodal}&	0.128&	0.162&	0.355&	1.249&0.392& 0.278& 0.571& 1.128\\
	MTN-TMT (Ours)& \textbf{0.142}& 	\textbf{0.171}&	\textbf{0.371}&	\textbf{1.357}&\textbf{0.402}&	\textbf{0.293}&	\textbf{0.587}&	\textbf{1.190} \\   \toprule
	\end{tabular}
        \label{avsd_summary}
	\end{table*}

\begin{table}[th]
	\setcounter{table}{3}
	\setlength{\belowcaptionskip}{-0.2cm}
	\centering
	\caption{Metrics of MTN-TMT with deeper TMT.}
	\setlength{\baselineskip}{1.5em}
	\setlength{\tabcolsep}{2mm}
	\normalsize
	\begin{tabular}{cccccc}
   \toprule
		\multicolumn{2}{c}{Depth}&\multirow{2}*{BLEU-4}&	\multirow{2}*{METEOR}&	\multirow{2}*{ROUGE-L}&	\multirow{2}*{CIDEr}\\
		 $M$& $N$	\\\cline{1-6} 
		\multicolumn{6}{c}{Text Only} \\\hline
		-&1&	0.346&	\textbf{0.289}&	\textbf{0.571}&	\textbf{1.107}\\ 
        		-&2&	0.343&	0.288&	0.568&	1.089\\
	 	-&3&	\textbf{0.349}&	0.288&	\textbf{0.571}&	1.093\\	 \hline
		\multicolumn{6}{c}{Video and Text} \\\hline
		1&1	&	0.393&	\textbf{0.296}&	\textbf{0.587}&	1.169\\ 
        		2&1	&	\textbf{0.402}&	0.293&	\textbf{0.587}&	\textbf{1.190}\\
	 	3&1	&	0.394&	0.288&	0.584&	1.167\\	 \cline{1-6}
	\toprule
	\end{tabular}
        \label{deeper-TMT}
	\end{table}

\begin{figure}[ht]
  \centering
  \setlength{\belowcaptionskip}{-0.2cm}
  \centerline{\includegraphics[scale=0.37]{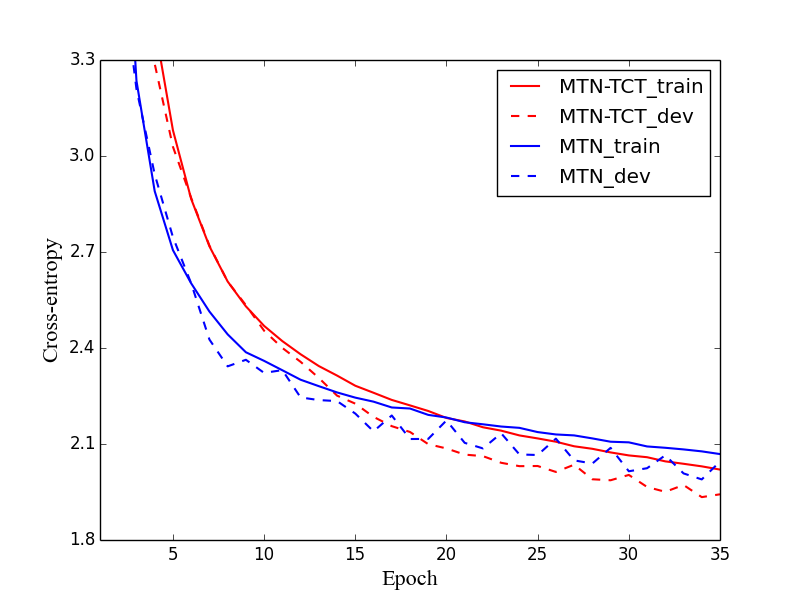}}
  \vspace{-0.5cm} 
  \centerline{}\medskip
\caption{Loss of MTN and MTN-TMT on the training set and dev set. Because dropout is activated when training but deactivated when evaluating on the dev set, training loss is slightly higher than dev loss.}
\label{loss_compare}
\end{figure}

\noindent
\textbf{Video and Text task.} Next, we evaluated MTN-TMT in the Video and Text task. The result is shown in the bottom part of Table \ref{avsd_result}. In this task, MTN-TMT improves upon the MTN by 0.004, 0.015, 0.016 and 0.062 on BLEU-4, METEOR, ROUGE-L, and CIDEr. Empirically, we find that MTN-TMT converges more stable to better results after 20 epochs when compared to MTN (see Figure \ref{loss_compare}). We speculate that such performance improvement occurs because TMT introduces better modal representations to MTN to accurately answer questions. \\

\noindent
\textbf{Ablation Study.} To further study the influence of the individual components in MTN-TMT, we perform comprehensive ablation analysis in the Video and Text task. First, we evaluate the performance of video-caption translator and dialog-summary translator, respectively. The results are shown in the lower part of Table \ref{avsd_result}. We observe that dialog-summary translator and video-caption translator are beneficial to improve the performance of MTN.  In addition, dialog-summary translator help MTN obtain the more competitive scores. 

Next, we consider supervised learning as a crucial factor for TMT. Consequently, we conduct experiments on decaying the learning weights $\alpha$ and $\beta$ simultaneously while training MTN-TMT. Note that we find the best hyperparameters of $\alpha$ and $\beta$ by grid search method on dev set. In this ablation study, we initialize $\alpha$ and $\beta$ to 0.3 and 0.8, respectively. Then, we decay the $\alpha$ and $\beta$ by $10\%$ every 10 epochs. When epoch reaches up to $20$, TMT can be regarded as a MulT-like method due to losing the supervision of caption and summary. As shown in Table \ref{decay_weight}, training TMT by supervised learning is a positive way to improve the representations of the modalities.

Considering deep TMT naturally integrates multiple-level features and would provide better representations for modality \cite{10.1007/978-3-319-10590-1_53}, we study the influence on depth of TMT. First, we increase the depth of TMT in Text Only task. The upper part of Table \ref{deeper-TMT} shows the results. However, deeper dialog-summary translator does not improve the performance in Text Only task. We suspect that TMT with one layer has sufficient ability to learn representations of dialog modality. Next, we keep the depth of dialog-summary translator to 1 and evaluate deeper video-caption translator on Video and Text task. The lower part of the Table \ref{deeper-TMT} shows results. We find that video-caption translator with 2 layers outperforms the other two, for example, improving CIDEr by 0.021. \\

\noindent
\textbf{Comparison with Submission Models.}
Finally, we summarize the performance of the recently proposed models in Text Only task and Video and Text task. The hierarchical attention (HAN) \cite{sanabria2019cmu} is  the systems that rank $1^{st}$ in the DSTC7-AVSD task adopted. Multimodal Transformer Networks (MTN) \cite{le2019multimodal} is the state-of-the-art system before the DSTC8-AVSD Challenge. As the DSTC7 dataset is organized as two versions (single reference and six references), we report the model performance on both sets. As shown in Table \ref{avsd_summary}, MTN-TMT surpasses MTN and other submission models in both Text Only task and Video and Text task. Compared with MTN, MTN-TMT obtains improvements on all metrics, especially, achieving relative improvement up to $14.1\%$ on CIDEr.

\section{Conclusion}
In this paper, we propose a Transformer-based Modal Translator (TMT) to learn the representations of multimodal sequence for Audio Visual Scene-aware Dialog. TMT enhance representations of the source modal sequence by translating the source modal sequence to the related target modal sequence. Based on MTN, we apply TMT to video and dialog to learn better representations of modalities in video-grounded dialog systems. On the AVSD track of the $7^{th}$ Dialog System Technology Challenge, MTN-TMT outperforms the MTN and other submission models in both Video and Text task and Text Only task. Compared with MTN, MTN-TMT improves all metrics, especially, achieving relative improvement up to $14.1\%$ on CIDEr.
\label{sec:conclusion}

\bibliographystyle{IEEEtran}

\bibliography{mybib}


\end{document}